\pdfoutput=1
\documentclass[11pt,a4paper,twocolumn,table,dvipsnames]{article}

\usepackage[final]{acl}

\usepackage[normalem]{ulem}
\usepackage{adjustbox}
\usepackage{algorithm}
\usepackage{algorithmic}
\usepackage{amsmath}
\usepackage{amssymb}
\usepackage{array}
\usepackage{balance}
\usepackage{blindtext}
\usepackage{booktabs}
\usepackage{caption}
\usepackage{colortbl}
\usepackage{enumerate}
\usepackage{float}
\usepackage{geometry}
\usepackage{hyperref}
\usepackage{latexsym}
\usepackage{listings}
\usepackage{makecell}
\usepackage{marvosym}
\usepackage[nopatch=footnote]{microtype} % solve the warning
\usepackage{multirow}
\usepackage{multicol}
\usepackage{pgf}
\usepackage{setspace}
\usepackage{stfloats}
\usepackage{subcaption}
\usepackage{svg}
\usepackage{tabularx}
\usepackage{tcolorbox}
\usepackage{times}
\usepackage{url}
\usepackage{color-edits}
\addauthor{xinyu}{cyan}

\definecolor{cellbase}{RGB}{245, 250, 245}    
\definecolor{cellmid}{RGB}{220, 240, 220}     
\definecolor{cellgood}{RGB}{190, 225, 190}    
\definecolor{cellbest}{RGB}{150, 210, 150}    

\newcommand{\cellcolorbyimprove}[2]{
  \pgfmathsetmacro{\improve}{(#2-#1)/max(1,#1)*100}
  \ifdim \improve pt > 40pt
    \cellcolor{cellbest} #2
  \else\ifdim \improve pt > 20pt
    \cellcolor{cellgood} #2
  \else\ifdim \improve pt > 10pt
    \cellcolor{cellmid} #2
  \else
    \cellcolor{cellbase} #2
  \fi\fi\fi
}

\newcommand{\cellcolorbyimprovenew}[2]{
  \pgfmathsetmacro{\improve}{(#2-#1)/max(1,#1)*100}
  \ifdim \improve pt > 20pt
    \cellcolor{cellbest} #2
  \else\ifdim \improve pt > 10pt
    \cellcolor{cellgood} #2
  \else\ifdim \improve pt > 5pt
    \cellcolor{cellmid} #2
  \else
    \cellcolor{cellbase} #2
  \fi\fi\fi
}

\newcommand{\Sref}[1]{\hyperref[#1]{\S\ref{#1}}}

\usepackage[T1]{fontenc}
\usepackage[utf8]{inputenc}

\usepackage{inconsolata}

\usepackage{graphicx}
\usepackage{cuted} 

\usepackage{soul}

\title{NOVER: Incentive Training for Language Models \\ via Verifier-Free Reinforcement Learning}

\author{
 \textbf{Wei Liu\textsuperscript{\footnotesize $^\bigstar$}},
 \textbf{Siya Qi\textsuperscript{\footnotesize $^\bigstar$}},
 \textbf{Xinyu Wang\textsuperscript{\footnotesize $^\bigstar$}},
 \textbf{Chen Qian\textsuperscript{\footnotesize $^\clubsuit$}},
 \textbf{Yali Du\textsuperscript{\footnotesize $^{\bigstar\spadesuit}$}},
 \textbf{Yulan He\textsuperscript{\footnotesize $^{\bigstar\spadesuit}$\textsuperscript{\Letter}}},
\\
\\
 \textsuperscript{\footnotesize $^\bigstar$}King's College London,
 \textsuperscript{\footnotesize $^\spadesuit$}The Alan Turing Institute,
 \textsuperscript{\footnotesize $^\clubsuit$}Shanghai Jiao Tong University
\\
 \small{
  \texttt{\{wei.4.liu, yali.du, yulan.he\}@kcl.ac.uk}
 }
}

\begin{document}
\maketitle

\begin{strip}
\small
\vspace{-2cm}
\begin{center}
\url{https://huggingface.co/spaces/thinkwee/NOVER}
\end{center}
\end{strip}

\begin{abstract}
Recent advances, such as DeepSeek R1-Zero, highlight the effectiveness of incentive training, a reinforcement learning paradigm that computes rewards solely based on the final answer part of a language model’s output, thereby encouraging the generation of intermediate reasoning steps. However, these methods fundamentally rely on external verifiers, which limits their applicability to domains like mathematics and coding, where such verifiers are readily available. Although reward models can serve as verifiers, they require high-quality annotated data and are costly to train.
In this work, we propose \textbf{NOVER}, \underline{NO}-\underline{VER}ifier Reinforcement Learning, a general reinforcement learning framework that requires only standard supervised fine-tuning data with no need for an external verifier. NOVER enables incentive training across a wide range of text-to-text tasks and outperforms the model of the same size distilled from large reasoning models such as DeepSeek R1 671B by 7.7\%. Moreover, the flexibility of NOVER enables new possibilities for optimizing large language models, such as inverse incentive training.
\end{abstract}

\section{Introduction}

\begin{figure}[htbp]
    \centering
    \includegraphics[width=0.99\linewidth]{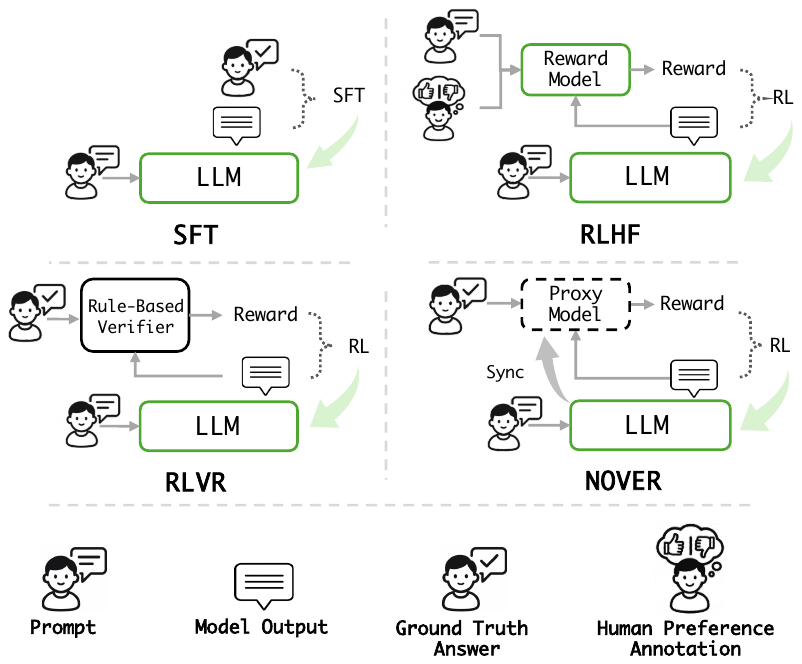}
    \caption{Comparison of NOVER with other post-training paradigms.}
    \label{fig:paradigm}
    \vspace{-0.5cm}
\end{figure}

\begin{figure*}[htbp]
    \centering
    \includegraphics[width=0.999\linewidth]{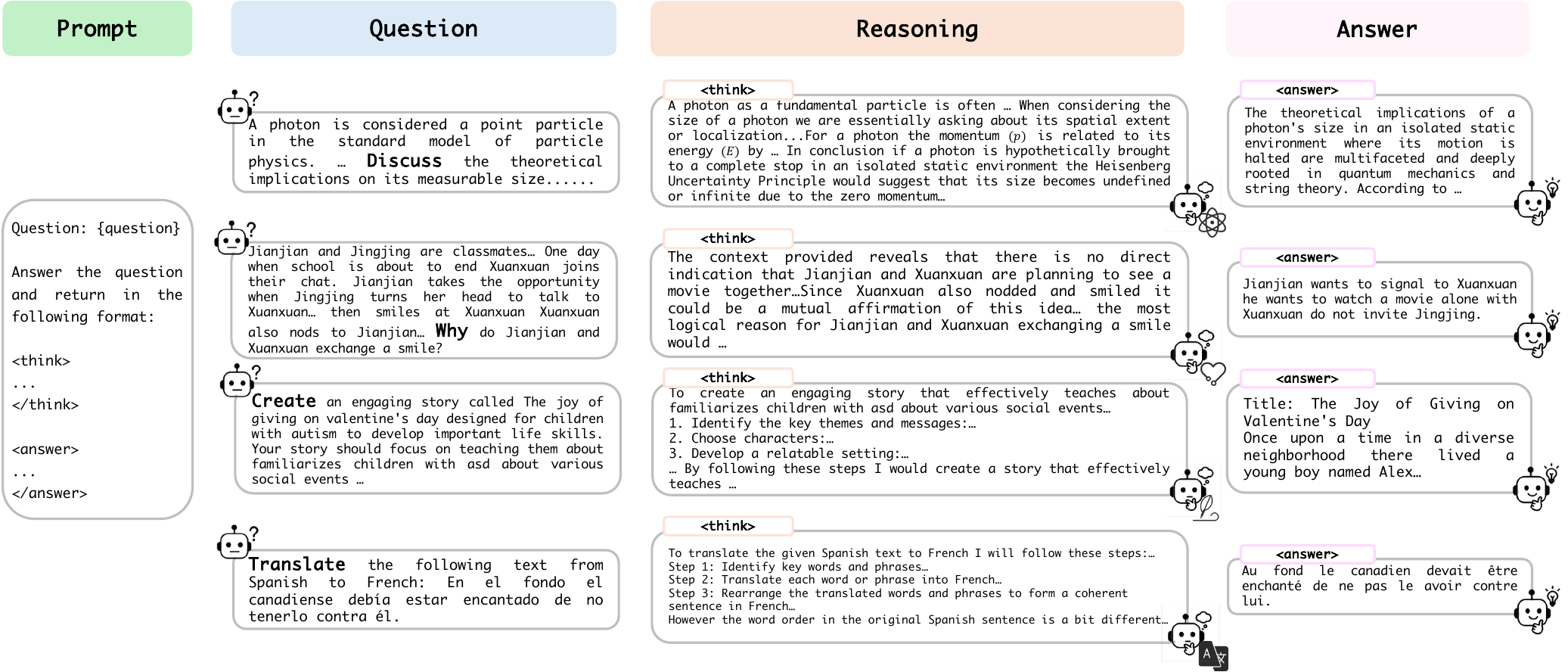}
    \caption{Examples of Qwen2.5-7B-NOVER on a range of text-to-text tasks, demonstrating its ability to handle open-ended questions such as “Discuss”, “Why”, or “Create”. These tasks often require free-form answers that are difficult to verify with clear-cut labels, posing challenges for incentive training.}
    \label{fig:case}
\end{figure*}

Recent progress in Large Language Model (LLM) reasoning has been accelerated by \emph{incentive training} \cite{guo2025deepseek, xie2025logic, yu2025dapo, zeng2025simplerl, liu2025understanding, yuan2025vapo, hu2025open}, a new Reinforcement Learning (RL) paradigm which optimizes models by computing rewards only on the final answer part in model response, and incentivize models to generate intermediate tokens like reasoning steps spontaneously. Notably, methods such as DeepSeek R1-Zero \cite{guo2025deepseek} have shown that in domains like mathematics and coding, using a simple rule-based verifier to calculate reward and perform RL-only incentive training can achieve impressive performance, known as \underline{R}einforcement \underline{L}earning with \underline{V}erifiable \underline{R}eward (RLVR). It has been demonstrated that such incentivized reasoning ability can generalize to areas like multi-modality \cite{shen2025vlm, zhang2025r1, yang2025r1, feng2025video} or language agent \cite{xia2025gui, jin2025search, song2025r1, wang2025ragen, feng2025retool}.

However, the success of incentive training with RLVR hinges on external verifiers that can judge the correctness of model outputs. For many highly contextualized and hard-to-grade tasks requiring similar reasoning or cognitive capability, ranging from social behavior analysis to creative writing, such verifiers are infeasible or hard to construct \cite{weng2025think}. Recent works have explored training general-purpose large verifier models across diverse domains \cite{generalreasoner, seed2025seed, su2025crossing}. However, building accurate verifier models involves a complex pipeline, which is significantly more costly than traditional reward models for preference alignment \cite{liu2025inference}. Moreover, once trained, these verifier models demand substantial computational resources to be deployed during the subsequent RL training.

To address these limitations of RLVR, we propose \textbf{NOVER}, \underline{NO}-\underline{VER}ifier Reinforcement Learning, a novel framework for incentive training without an external verifier. As shown in \autoref{fig:paradigm}, compared with traditional Supervised Fine-Tuning (SFT), NOVER can perform incentive training similar to RLVR; compared to Reinforcement Learning from Human Feedback (RLHF), NOVER does not need a trained reward model for verification; and compared with RLVR, NOVER extends the incentive training to areas where a reliable verifier is hard to acquire. NOVER requires only standard SFT data and utilizes the model itself to build a reward proxy for lightweight RL training. By calculating perplexity-based reward based on the model's reasoning process, it enables incentive-driven reinforcement learning across a wide range of text-to-text tasks, as shown in \autoref{fig:case}. 
Our main contributions are as follows: 
\begin{enumerate}
    \item We introduce \textbf{NOVER}, a verifier-free incentive training framework that generalizes RLVR to arbitrary text-to-text tasks with minimal SFT data. 
    \item Our experiments and analysis demonstrate NOVER's superior performance. It is stable to train compared with model-as-verifier methods and mitigates reward hacking, shapes reasoning patterns, and adapts to diverse tasks.
    \item NOVER’s flexibility enables imaginative approaches such as inverse incentive training. This paradigm teaches a model how to fish rather than simply giving it a fish, surpassing standard incentive training on tasks that need creativity.
\end{enumerate}

\section{Related Work}

\paragraph{Language Model Reasoning} 
Early research designed prompting techniques and workflows based on human cognitive priors to enhance the reasoning capabilities of LLMs. Chain-of-Thought (CoT) prompting \cite{wei2022chain} enabled step-by-step reasoning, later extended by Zero-Shot CoT \cite{kojima2022large} and Self-Consistency \cite{wang2022self}. More recent methods, such as Tree of Thoughts \cite{yao2023tree}, Least-to-Most \cite{zhou2022least}, Plan-and-Solve \cite{wang2023plansolve}, Sketch-Navigation \cite{liuautonomous}, and Multi-Perspective Self-Reflection \cite{yan2024mirror}, introduced structured exploration over reasoning trajectories. 
ReAct \cite{yao2022react} combined reasoning with external tool use to enable more interactive problem-solving. 
Beyond prompting, verification-based approaches were proposed \cite{cobbe2021training} to solve math reasoning. Neuro-symbolic methods \cite{pan2023logic} fuse LLMs with symbolic solvers. Previous works also utilize Process Reward Models \cite{lightman2023let} combined with Monte Carlo Tree Search \cite{kocsis2006bandit} for step-level exploration to replicate OpenAI's o1 \cite{jaech2024openai}.

\paragraph{Incentive Training}
In contrast to the above methods, DeepSeek-R1 \cite{guo2025deepseek} proposed a simpler paradigm called incentive training. Building on this idea, recent works such as Logic-RL \cite{xie2025logic}, DAPO \cite{yu2025dapo}, SimpleRL \cite{zeng2025simplerl}, OpenReasoner \cite{hu2025open}, Dr.GRPO \cite{liu2025understanding}, and VAPO \cite{yuan2025vapo} investigated best practices on aspects such as exploration-exploitation, emergence of "aha" moments, and task difficulty variance. 
Several recent efforts attempted to extend incentive training beyond math and coding \cite{he2025r1, lu2025tom, gurung2025learning, su2025crossing, generalreasoner}. However, these approaches often rely on domain-specific rules for verification or involve training large verifier models, which can be computationally expensive.

\section{Background}
\paragraph{Rule-based Reward}
Given a training prompt $p$ with a template \cite{guo2025deepseek} asking to generate intermediate tokens $t$ (e.g., reasoning steps) followed by final answer $a$, incentive training aims to use RL to optimize a model $\pi_{\theta}$ with outcome reward $R_{\mathrm{rule}}$, which is solely computed by a rule-based verifier $v: \mathcal{A} \to \{0,1\}$ on $a$:
\begin{equation}\label{eq:rule_reward}
R_{\mathrm{rule}} = v(a)
\end{equation}
Such a sparse but accurate outcome reward encourages the model to autonomously generate intermediate tokens $t$ that could lead to a better $a$. In reasoning tasks, for example, $t$ may be reasoning behaviors like task decomposition or self-reflection.
To ensure the verifier can easily parse outputs, DeepSeek-R1 introduces a tag format reward \( R_{\mathrm{f}} \), which requires \( t \) to be enclosed in \texttt{<think>} tags and \( a \) in \texttt{<answer>} tags:
\begin{equation}\label{eq:tag_reward}
R_{\mathrm{f}}= f_{\mathrm{format}}(t, a)
\end{equation}
where $f_{\mathrm{format}} : \mathcal{T} \times \mathcal{A} \to \{0,1\}$ is the regular expression based format check function. Then the combined reward is:
\begin{equation}
R(p, t, a) = w_{\mathrm{rule}} \cdot R_{\mathrm{rule}} + w_{\mathrm{f}} \cdot R_{\mathrm{f}}
\end{equation}

\paragraph{GRPO}
After the calculation of reward, Group Relative Policy Optimization (GRPO) \cite{shao2024deepseekmath} is then used to optimize $\pi_{\theta}$. For each prompt $p$, the model rolls out a group of completions $\mathcal{C} = \{(t_1, a_1), ..., (t_G, a_G)\}$. The group-normalized advantage is:
\begin{equation}
A_i = \frac{R(p, t_i, a_i) - \mu_{\mathcal{C}}}{\sigma_{\mathcal{C}}}
\end{equation}
where $\mu_{\mathcal{C}}$ and $\sigma_{\mathcal{C}}$ are the mean and standard deviation of rewards in the group.
Then the policy is updated with a clipped objective:

\vspace{-0.5\baselineskip}

\begingroup
\footnotesize
\begin{equation}
\mathcal{J}_{\text{GRPO}}(\theta) = 
\mathbb{E}_{\{p, t_i, a_i\}}\left( \frac{1}{G} \sum_{i=1}^G r_i^{clip} - \beta D_{\text{KL}}(\pi_\theta \,\|\, \pi_{\text{ref}})\right)
\end{equation}
\begin{equation}
r_i^{clip} = \min \left( r_i, \operatorname{clip}(r_i, 1{-}\epsilon, 1{+}\epsilon) \right) A_i
\end{equation}
\begin{equation}
r_i = \frac{\pi_\theta(t_i, a_i \mid p)}{\pi_{\theta_{\text{old}}}(t_i, a_i \mid p)}
\end{equation}
\endgroup
\noindent where $\epsilon$ is the clip ratio and $\beta$ is the KL penalty weight. $r$ is the policy ratio \cite{schulman2017proximal}.
This framework enables LLMs to develop explicit, high-quality reasoning processes without dense supervision for intermediate steps.

\section{Method}

\begin{figure}[htbp]
    \centering
    \includegraphics[width=0.999\linewidth]{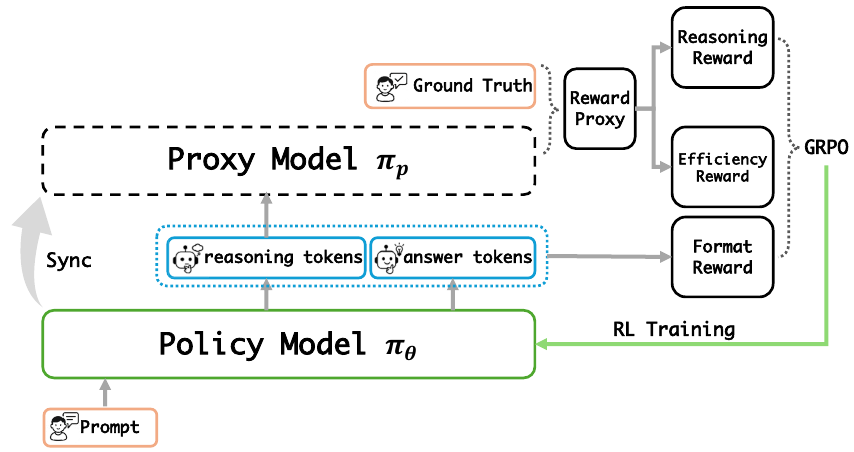}
    \caption{The overall process of NOVER.}
    \label{fig:overall}
\end{figure}

It can be seen in \autoref{eq:rule_reward} that the incentive training described above requires a verifier to compute $R_{\mathrm{rule}}$. \textbf{NOVER} eliminates the need for a verifier by introducing reasoning perplexity as a reward proxy, which can be calculated on any SFT data.

\subsection{Reasoning Perplexity as Proxy}

The core idea of NOVER is that the perplexity of the ground truth answer, conditioned on the model's reasoning trajectory, can serve as a natural proxy for reward. Given a prompt $p$, the ground truth answer $g$, and a model response consisting of reasoning tokens $t$ and answer tokens $a$, we compute the reasoning perplexity $P_r$ using a proxy model $\pi_{\mathrm{p}}$ as follows:

\vspace{-0.5\baselineskip}

\begingroup
\small
\begin{equation}
P_r(p, t, g) = \exp\left(-\frac{\sum_{i=1}^{|g|} \log \pi_{\mathrm{p}}(g_i \mid p, t, g_{<i})}{|g| \cdot N(|t|)}\right)
\end{equation}
\endgroup
\noindent where $|g|$ denotes the number of tokens in the ground truth. $N(|t|)$ is a simple normalization factor based on the length of the reasoning tokens, which alleviates the length bias of perplexity:
\begin{equation}
N(|t|) = \max\left(1, 1 + \log(|t|)\right)
\end{equation}
A lower reasoning perplexity $P_r$ indicates a higher probability of generating the correct answer based on the reasoning tokens. As the calculation of $P_r$ is done in a teacher-forcing way without any decoding, computation of $P_r$ requires only $\sim 5\%$ of total training time (see details in \autoref{appendix:profiling}. Then $P_r$ can be utilized to calculate the reward defined in \Sref{reward_define}.

\subsection{Policy-Proxy Synchronization}

Unlike recent work that employs an extra frozen model as the proxy \cite{gurung2025learning}, we instead use the policy model $\pi_{\theta}$ itself as the proxy $\pi_{\mathrm{p}}$. This choice reflects their shared goal: minimizing the perplexity of correct answers given high-quality reasoning. Moreover, an extra frozen proxy can diverge from the evolving policy, leading to inconsistency (see \Sref{sec:curse}). 
Importantly, using $\pi_{\theta}$ as $\pi_{\mathrm{p}}$ does not mean the model acts as both athlete and referee, as $\pi_{\mathrm{p}}$ itself does not judge but leverages the objective ground truth $g$ for fair evaluation. 
In practice, we initialize both $\pi_{\theta}$ and $\pi_{\mathrm{p}}$ from the same pretrained checkpoint and periodically sync $\pi_{\mathrm{\theta}}$ to $\pi_{p}$ every $T_{\mathrm{sync}}$ steps via exponential smoothing, following TR-DPO \cite{gorbatovski2024learn}.

\vspace{-1.0\baselineskip}

\begin{equation}
\pi_{\mathrm{p}} \leftarrow \alpha \cdot \pi_{\mathrm{p}} + (1-\alpha) \cdot \pi_{\theta}
\end{equation}

where $\alpha \in [0,1]$ denotes the synchronization rate. Such synchronization enables the proxy model to gradually adapt to improvements in the policy while ensuring a stable calculation of perplexity. Since we employ LoRA~\cite{hu2022lora} for efficient training, the $\pi_{\theta}$ and $\pi_{\mathrm{p}}$ share the same base model and can be switched seamlessly by replacing the LoRA adapter, which comprises only about 0.4\% of the full model parameters.

\subsection{Verifier-Free Reward}\label{reward_define}

Given a prompt $p$ and its corresponding group of completions $(t_i, a_i)$, we compute two rewards based on the reasoning perplexity $P_r$.

\paragraph{Reasoning Reward} Reasoning perplexity $P_r$ can serve directly as a reward, as it reflects the quality of the reasoning trajectory. VR-CLI \cite{gurung2025learning} and reasoning advantage SFT \cite{foster2024reward} suggest calculating relative perplexity gain to stabilize training, but we found that GRPO’s group normalization already achieves a similar calculation, so there is no need for such a redundant operation. In practice, we observe that relative perplexity improvements still vary widely across different samples, causing high variance in clip ratios and unstable training. To mitigate this, we propose to discretize reasoning perplexity into reasoning rewards $R_r$. For each completion $(p_i, t_i, a_i)$ with $P_r^i$, we compute $R_r$ based on its rank among $n_{\mathrm{valid}}$ valid completions. Let $\mathcal{P} = \{P_r^1, P_r^2, ..., P_r^{n_{\mathrm{valid}}}\}$, sorted as $P_r^{(1)} \leq \cdots \leq P_r^{(n_{\mathrm{valid}})}$, reasoning rewards are assigned by quantile rank:

\vspace{-0.5\baselineskip}

\begingroup
\small
\begin{equation}
R_r(p_i, t_i, a_i) = 
\begin{cases}
\frac{n_{\mathrm{valid}} - \mathrm{rank}(P_r^i) + 1}{n_{\mathrm{valid}}}, & \text{if } \mathrm{rank}(P_r^i) \leq k \\
0, & \text{otherwise}
\end{cases}
\end{equation}
\endgroup

In practice, we use either $k = 1$ or $k = n_{\mathrm{valid}}$. The former assigns a reward of 1.0 only to the best (lowest-perplexity) completion, which is suitable for tasks with objectively correct answers. The latter distributes rewards to all completions based on their normalized rank and is better suited for subjective or open-ended tasks.

\paragraph{Efficiency Reward} Recent work \cite{sui2025stop} suggests that longer reasoning is not always better, and effective reasoning length is what matters. We introduce an Efficiency Reward $R_{\mathrm{e}}$, encouraging the model to generate shorter but better reasoning trajectories. A completion should be rewarded if it achieves lower reasoning perplexity with fewer reasoning tokens compared to other completions. The $R_{\mathrm{e}}$ can be calculated as:

\begin{small}
\begin{equation}
R_{\mathrm{e}}(p_i, t_i, a_i) = \frac{\sum_{\substack{j=1, j \neq i}}^{n} \mathbf{1}\left[P_r^i < P_r^j \land |t_i| < |t_j|\right]}{n_{\mathrm{valid}} - 1} 
\end{equation}
\end{small}

It is notable that although $R_{\mathrm{e}}$ shares a similar target with $R_r$, where lower perplexity is better, we cannot use $R_{\mathrm{e}}$ to replace $R_r$ since in the early training stage, most completions could not develop efficient reasoning to acquire the $R_{\mathrm{e}}$.

These two rewards are then combined with the tag format reward $R_{\mathrm{f}}$ in \autoref{eq:tag_reward} to form the final reward for GRPO training. The combined reward function is then:
\begin{equation}
R_{\mathrm{total}} = w_{\mathrm{f}} R_{\mathrm{f}} + \mathbb{I}(R_{\mathrm{f}} = 1) \cdot (w_{\mathrm{r}} R_{\mathrm{r}} + w_{\mathrm{e}} R_{\mathrm{e}})
\end{equation}
where $w_{\mathrm{f}}, w_{\mathrm{r}}, w_{\mathrm{e}}$ are weights that control the relative importance of each reward component. All these rewards are in the range $[0,1]$. Notably, they are conditionally dependent rather than simply additive: only when the tag format is correct ($R_{\mathrm{f}}= 1$) are the other two rewards calculated and added; otherwise, they are set to 0. We provide a detailed analysis of how this design effectively prevents the "curse of proxy" in \Sref{sec:curse}. 
NOVER can be combined with various RL algorithms, such as PPO \cite{schulman2017proximal}, and here we use GRPO for efficient training.

\section{Experimental Setup}

\paragraph{Dataset} To evaluate the effectiveness of the NOVER-incentivized model, we select datasets spanning four broad domains: \ul{(1) General Reasoning}: This category includes challenging tasks that go beyond standard STEM benchmarks, requiring complex %, rationale-based 
reasoning, including three datasets, \textbf{Natural Reasoning} \citep{yuan2025naturalreasoning}, \textbf{General Thought} \citep{generalthought430k}, and \textbf{WebInstruct} \citep{webinstruct_verified}, which require models to produce factually grounded answers, often involving multi-step reasoning and justification. \ul{(2) Creative Writing}: To assess long-form narrative generation and planning ability, we use \textbf{SS-GEN} \citep{feng2024ssgensocialstorygeneration}, a benchmark focused on coherent, socially themed storytelling. \ul{(3) Social Intelligence}: We evaluate models on emotionally and socially grounded reasoning using \textbf{EmoBench} \citep{sabour-etal-2024-emobench} and \textbf{ToMBench} \citep{chen2024tombench}, which test emotion recognition, social action prediction, and theory-of-mind reasoning. \ul{(4) Multilingual Ability}: We employ the \textbf{OPUS book} corpus \citep{tiedemann-2012-parallel}, which includes translation tasks across 16 languages and 64 source-target language pairs, to evaluate cross-lingual reasoning and generalization. For all datasets and subcategories, we construct training and test sets via uniform sampling. Further details are in \autoref{appendix:dataset}.

{
\setlength{\tabcolsep}{4pt}
\begin{table*}[htbp]
\centering
\fontsize{7}{9}\selectfont
\begin{tabular}{l|ccc|c|cc|c|c}
\toprule
\multirow{2}{*}{\centering\textbf{Method}} 
& \multicolumn{3}{c|}{\textbf{General Reasoning}} 
& \multicolumn{1}{c|}{\textbf{Writing}} 
& \multicolumn{2}{c|}{\textbf{Social Intelligence}} 
& \textbf{Multilingual} 
& \multirow{2}{*}{\textbf{Avg.}} \\
\cmidrule(lr){2-4} \cmidrule(lr){5-5} \cmidrule(lr){6-7} \cmidrule(lr){8-8}
& NR & GT & WI 
& SGN 
& EB & TB
& OPUS 
&  \\
\midrule
\multicolumn{9}{l}{\textbf{Qwen2.5-3B}} \\
$Base\ Model$      
& \cellcolorbyimprove{21.80}{21.80}\% & \cellcolorbyimprove{43.10}{43.10}\% & \cellcolorbyimprove{18.40}{18.40}\% 
 & \cellcolorbyimprove{18.70}{18.70}\% 
& \cellcolorbyimprove{32.03}{32.03}\% & \cellcolorbyimprove{46.79}{46.79}\%
& \cellcolorbyimprove{16.70}{16.70}\% 
& \cellcolorbyimprove{28.22}{28.22}\% \\
$+\ CoT$      
& \cellcolorbyimprove{21.80}{24.40}\% & \cellcolorbyimprove{43.10}{48.90}\% & \cellcolorbyimprove{18.40}{24.20}\% 
 & \cellcolorbyimprove{18.70}{14.76}\% 
& \cellcolorbyimprove{32.03}{28.12}\% & \cellcolorbyimprove{46.79}{51.23}\%
& \cellcolorbyimprove{16.70}{1.40}\% 
& \cellcolorbyimprove{28.22}{27.57}\% \\
$+\ SFT$             
& \cellcolorbyimprove{21.80}{27.00}\% & \cellcolorbyimprove{43.10}{36.20}\% & \cellcolorbyimprove{18.40}{27.30}\% 
 & \cellcolorbyimprove{18.70}{20.08}\% 
& \cellcolorbyimprove{32.03}{36.72}\% & \cellcolorbyimprove{46.79}{48.66}\%
& \cellcolorbyimprove{16.70}{17.30}\% 
& \cellcolorbyimprove{28.22}{30.47}\% \\
\textbf{$+\ NOVER$} 
& \textbf{\cellcolorbyimprove{21.80}{28.60}\%} & \textbf{\cellcolorbyimprove{43.10}{60.30}\%} & \textbf{\cellcolorbyimprove{18.40}{28.10}\%} 
 & \textbf{\cellcolorbyimprove{18.70}{41.64}\%} 
& \textbf{\cellcolorbyimprove{32.03}{38.28}\%} & \textbf{\cellcolorbyimprove{46.79}{57.88}\%}
& \textbf{\cellcolorbyimprove{16.70}{20.70}\%} 
& \textbf{\cellcolorbyimprove{28.22}{39.36}\%} \\
\midrule
\multicolumn{9}{l}{\textbf{Qwen2.5-7B}} \\
$Base\ Model$      
& \cellcolorbyimprove{31.80}{31.80}\% & \cellcolorbyimprove{48.50}{48.50}\% & \cellcolorbyimprove{20.70}{20.70}\% 
 & \cellcolorbyimprove{24.21}{24.21}\% 
& \cellcolorbyimprove{28.91}{28.91}\% & \cellcolorbyimprove{44.22}{44.22}\%
& \cellcolorbyimprove{19.30}{19.30}\% 
& \cellcolorbyimprove{31.09}{31.09}\% \\
$+\ CoT$     
& \cellcolorbyimprove{31.80}{31.20}\% & \cellcolorbyimprove{48.50}{57.60}\% & \cellcolorbyimprove{20.70}{29.20}\% 
 & \cellcolorbyimprove{24.21}{33.46}\% 
& \cellcolorbyimprove{28.91}{38.28}\% & \cellcolorbyimprove{44.22}{50.99}\%
& \cellcolorbyimprove{19.30}{1.60}\% 
& \cellcolorbyimprove{31.09}{34.62}\% \\
$+\ SFT$             
& \cellcolorbyimprove{31.80}{27.50}\% & \cellcolorbyimprove{48.50}{45.20}\% & \cellcolorbyimprove{20.70}{33.50}\% 
 & \cellcolorbyimprove{24.21}{37.85}\% 
& \cellcolorbyimprove{28.91}{47.66}\% & \cellcolorbyimprove{44.22}{57.06}\%
& \cellcolorbyimprove{19.30}{23.30}\% 
& \cellcolorbyimprove{31.09}{38.87}\% \\
\textbf{$+\ NOVER$} 
& \textbf{\cellcolorbyimprove{31.80}{38.20}\%} & \textbf{\cellcolorbyimprove{48.50}{61.80}\%} & \textbf{\cellcolorbyimprove{20.70}{36.60}\%} 
 & \textbf{\cellcolorbyimprove{24.21}{50.79}\%} 
& \textbf{\cellcolorbyimprove{28.91}{49.22}\%} & \textbf{\cellcolorbyimprove{44.22}{67.79}\%}
& \textbf{\cellcolorbyimprove{19.30}{26.80}\%} 
& \textbf{\cellcolorbyimprove{31.09}{47.31}\%} \\
\midrule
Qwen2.5-3B-Instruct    
& \cellcolorbyimprove{21.80}{27.10}\% & \cellcolorbyimprove{43.10}{50.00}\% & \cellcolorbyimprove{18.40}{31.50}\% 
 & \cellcolorbyimprove{18.70}{21.25}\% 
& \cellcolorbyimprove{32.03}{40.62}\% & \cellcolorbyimprove{46.79}{58.69}\%
& \cellcolorbyimprove{16.70}{19.90}\% 
& \cellcolorbyimprove{28.22}{35.58}\% \\
Qwen2.5-7B-Instruct    
& \cellcolorbyimprove{31.80}{29.90}\% & \cellcolorbyimprove{48.50}{56.20}\% & \cellcolorbyimprove{20.70}{35.60}\% 
 & \cellcolorbyimprove{24.21}{67.72}\% 
& \cellcolorbyimprove{28.91}{46.88}\% & \cellcolorbyimprove{44.22}{65.23}\%
& \cellcolorbyimprove{19.30}{23.50}\% 
& \cellcolorbyimprove{31.09}{46.43}\% \\
R1-Distill-Qwen-7B      
& \cellcolorbyimprove{31.80}{41.00}\% & \cellcolorbyimprove{48.50}{60.20}\% & \cellcolorbyimprove{20.70}{38.00}\% 
 & \cellcolorbyimprove{24.21}{40.16}\% 
& \cellcolorbyimprove{28.91}{35.16}\% & \cellcolorbyimprove{44.22}{54.61}\%
& \cellcolorbyimprove{19.30}{8.20}\% 
& \cellcolorbyimprove{31.09}{39.62}\% \\
\bottomrule
\end{tabular}
\caption{Overall performance. Each cell is shaded based on its relative improvement. Values in \textbf{bold} indicate the best-performing variant, excluding other post-trained models. \textbf{NR}: Natural Reasoning, \textbf{GT}: General Thoughts-430k, \textbf{WI}: WebInstruct, \textbf{SGN}: SS-GEN, \textbf{EB}: EmoBench, \textbf{TB}: TomBench, \textbf{OP}: OPUS-BOOK-TRANSLATION.}
\label{tab:main-results}
\end{table*}
}

\paragraph{Baselines}
We evaluate 3B and 7B versions of Qwen 2.5 \citep{qwen2.5} against several baselines. Prior work suggests that instruction-following and basic \textit{CoT} reasoning can emerge during multi-stage pretraining \cite{wei2022chain, zeng2025simplerl}. Thus, we include both vanilla and CoT responses to isolate the effect of NOVER incentive tuning from capabilities already acquired in pretraining. We also consider an SFT baseline, where the same pretrained checkpoints are fine-tuned on each dataset’s training split (identical to that used by NOVER). This comparison highlights whether NOVER enhances generalization and abstraction in contrast to SFT \cite{chu2025sft}. For reference, we report results from three strong post-trained models: Qwen2.5-3B/7B-Instruct, and R1-Distill-Qwen-7B, a 7B variant distilled from DeepSeek-R1 671B \citep{guo2025deepseek}. These models, trained with massive instruction-following and reasoning data, serve as high-performance baselines.

\paragraph{Training and Evaluation}
Our training framework is built on Huggingface TRL \cite{vonwerra2022trl}, using LoRA adapters \cite{hu2022lora} for efficient fine-tuning. Full hyperparameters are listed in \autoref{appendix:hyperparams}. Training time is mainly determined by rollout efficiency, which depends on the maximum generation length and varies by task (\autoref{appendix:dataset}). For example, training Qwen2.5-7B for 1,000 steps with a max length of 1,024 tokens on two H100 GPUs takes about 2 hours. For evaluation, we report accuracy on all datasets. For choice questions, we extend the extractor from \citet{pteam2025supergpqascalingllmevaluation} to identify both the option letter and corresponding text, then match against ground truth. For open-ended QA, we use Gemini-2.0-flash \cite{team2023gemini, team2024gemini} in an LLM-as-a-Judge setup (see details in \autoref{appendix:judge_prompt}).

\section{Results and Analysis}
We present a comprehensive and structured experiment organized into three main parts. \ul{(1) To examine \textit{when} NOVER performs well}, \Sref{sec:overall} analyzes its performance across different task types, and \Sref{sec:causes} investigates how data distribution influence reasoning ability. \ul{(2) To understand \textit{why} NOVER is effective}, \Sref{sec:curse} examines how NOVER addresses reward hacking for stable learning; \Sref{sec:compare_llm_judge} compares its verifier-free design with Model as Verifier approaches; and \Sref{sec:pattern} studies how incentivized reasoning patterns evolve during training. \ul{(3) To demonstrates NOVER’s \textit{flexibility}}, in \Sref{sec:rubric} we explored inverse incentive training for creative writing.

\begin{figure*}[htbp]
    \centering
    \includegraphics[width=0.99\linewidth]{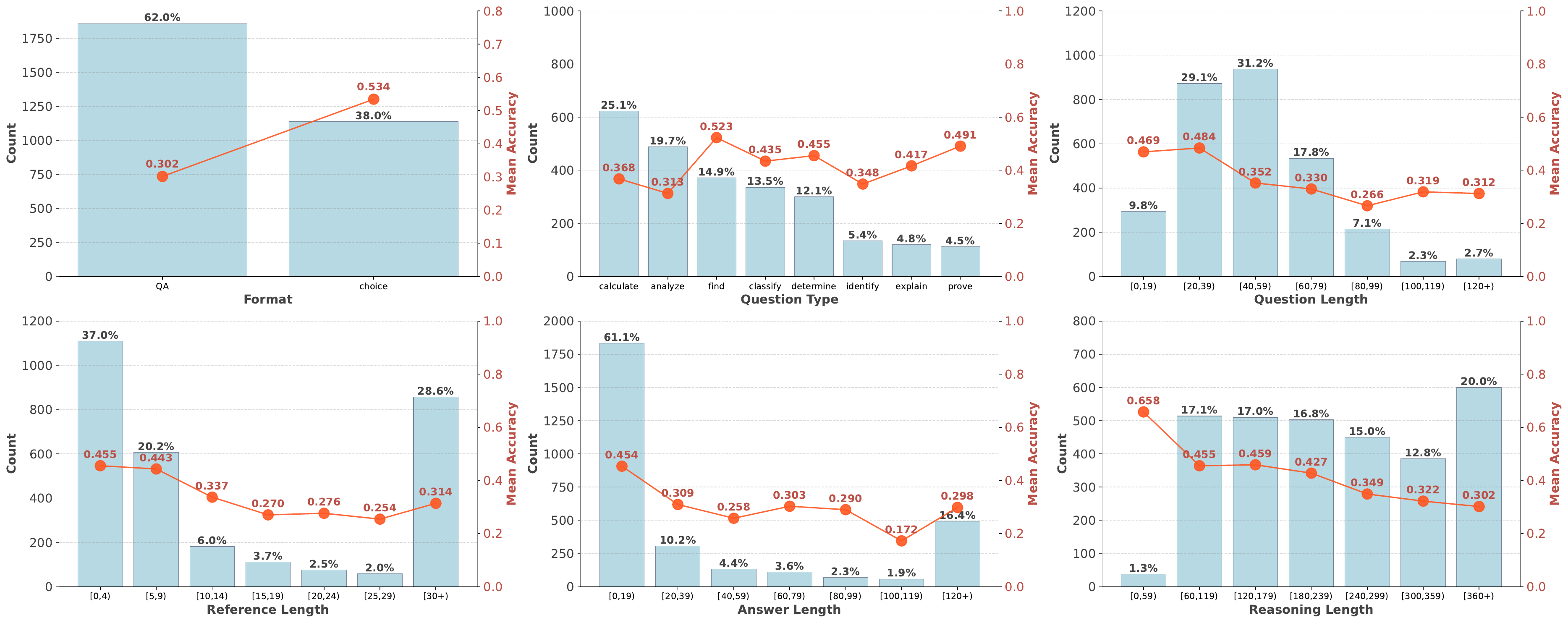}
    \caption{Accuracy of NOVER on three general reasoning tasks by the distribution of different aspects.}
    \label{fig:dis}
\end{figure*}

\subsection{Overall Performance}
\label{sec:overall}
\autoref{tab:main-results} summarizes the performance of NOVER and all baselines. We observe that, for both 3B and 7B model sizes, NOVER consistently outperforms all baselines across all types of tasks. Remarkably, NOVER enables the 3B model to achieve nearly 60\% accuracy on General Thoughts, approaching the performance of the much larger R1-Distill-Qwen-7B model distilled from a 671B teacher. In \autoref{appendix:RP_case}, we give a detailed example of how reasoning perplexity discriminates between good and bad responses, thus helping the model to reinforce learning. 
The improvement is particularly pronounced on datasets that challenge areas where less data is pretrained, such as EmoBench (for social intelligence) and OPUS (for multilingual). In these areas, direct CoT prompting may reduce accuracy. Our manual inspection of CoT outputs reveals that the base model is capable of generating well-structured and fluent CoT. However, these CoT-generated rationales can be easily hallucinated \cite{huang2025survey, ye2023cognitive, li2024deceptive}. In contrast, NOVER effectively corrects such hallucinated reasoning processes, as illustrated by a detailed example in \autoref{appendix:case}.
Notably, SFT sometimes even underperforms the base model, as it encourages the model to directly map questions to answers without an explicit reasoning process. In contrast, the CoT, NOVER, and even the base model can generate intermediate reasoning tokens that aid in answering.

\setlength{\tabcolsep}{4pt}
\begin{table}[htbp]
\centering
\fontsize{7}{9}\selectfont
\begin{tabular}{l|l|ccc}
\toprule
\textbf{Model} & \textbf{Method} & \textbf{NR} & \textbf{GT} & \textbf{WI} \\
\midrule
\multirow{4}{*}{\centering Llama-3.1-8B}
& $Base\ Model$ 
& \cellcolorbyimprovenew{34.20}{34.20}\% & \cellcolorbyimprovenew{36.70}{36.70}\% & \cellcolorbyimprovenew{29.90}{29.90}\% \\
& $+\ CoT$ 
& \cellcolorbyimprovenew{34.20}{28.10}\% & \cellcolorbyimprovenew{36.70}{35.10}\% & \cellcolorbyimprovenew{29.90}{30.00}\% \\
& $+\ SFT$ 
& \cellcolorbyimprovenew{34.20}{23.60}\% & \cellcolorbyimprovenew{36.70}{23.40}\% & \textbf{\cellcolorbyimprovenew{29.90}{34.50}\%} \\
& \textbf{$+\ NOVER$} 
& \textbf{\cellcolorbyimprovenew{34.20}{40.70}\%} & \textbf{\cellcolorbyimprovenew{36.70}{41.50}\%} & \cellcolorbyimprovenew{29.90}{34.00}\% \\
\midrule
\multirow{4}{*}{\centering Mistral-7B}
& $Base\ Model$ 
& \textbf{\cellcolorbyimprovenew{33.00}{33.00}\%} & \cellcolorbyimprovenew{17.80}{17.80}\% & \cellcolorbyimprovenew{27.00}{27.00}\% \\
& $+\ CoT$ 
& \cellcolorbyimprovenew{33.00}{29.20}\% & \cellcolorbyimprovenew{17.80}{18.60}\% & \cellcolorbyimprovenew{27.00}{27.10}\% \\
& $+\ SFT$ 
& \cellcolorbyimprovenew{33.00}{22.50}\% & \cellcolorbyimprovenew{17.80}{20.70}\% & \cellcolorbyimprovenew{27.00}{27.80}\% \\
& \textbf{$+\ NOVER$} 
& \cellcolorbyimprovenew{33.00}{32.20}\% & \textbf{\cellcolorbyimprovenew{17.80}{21.90}\%} & \textbf{\cellcolorbyimprovenew{27.00}{29.30}\%} \\
\bottomrule
\end{tabular}
\caption{General Reasoning performance with different instruct model backends.}
\label{tab:general-reasoning-backends}
\end{table}

In addition, as shown in \autoref{tab:general-reasoning-backends}, we also evaluated other model families such as Mistral~\cite{jiang2024mistral} and LLaMA~\cite{dubey2024llama}. Since the base checkpoint of Qwen exhibits strong instruction following ability, it can be directly optimized with reinforcement learning in an R1-Zero fashion. In contrast, the base checkpoints of other models lack such capability, so we employed instruct-tuned checkpoints instead. The results show that NOVER achieves consistent gains across different model backends and checkpoint types (pretrained or instruct-tuned).

\subsection{When and Where for Effective Reasoning}
\label{sec:causes}

\paragraph{Question format}We analyze NOVER's performance distribution on the general reasoning area, as shown in \autoref{fig:dis}. For question format, the accuracy on multiple-choice questions remains consistently higher than that on open-ended QA. This is primarily because the presence of candidate options in the question effectively reduces the search space during RL optimization, thereby lowering the task difficulty.
We further extract and analyze the key action words from the questions. The model achieves higher accuracy on questions with clear solution directions, such as \emph{find}, \emph{determine}, and \emph{classify}, since the corresponding reasoning chains are more likely to be sampled and reinforced during training. In contrast, for questions with more flexible requirements and less prior constraint on the reasoning process, such as \emph{analyze}, the model's performance is relatively weaker.

\paragraph{Through the lens of length}We also analyze the effect of length across four components: question, reference, generated answer, and reasoning process. For the first three, we observe that shorter inputs or outputs generally lead to higher accuracy. Interestingly, for the length of the model-generated reasoning process, accuracy remains relatively stable within the range of 60 to 240 tokens. This suggests that NOVER effectively incentivizes the model to adaptively generate reasoning of appropriate length according to the difficulty of each question, a property that is closely related to the design of the efficiency reward.

\begin{table}[htbp]
\centering
\small
\begin{tabular}{c|cc}
\toprule
    \textbf{Method}            & \textbf{3B} & \textbf{7B} \\ \midrule
$Base$   & 12.43\%             & 14.59\%             \\
$+\ CoT$   & 14.23\%             & 19.28\%             \\
$+\ SFT$   & 26.49\%             & 29.73\%             \\ \midrule
$+\ NOVER$ & 18.74\%             & 23.42\%             \\ \bottomrule
\end{tabular}
\caption{Experiments on FANToM \cite{kim2023fantom}, a theory of mind task with false premise problems.}
\label{tab:fantom}
\end{table}

\vspace{-0.5\baselineskip}

\paragraph{False Premise Task} Recent work on RLVR suggests that RL may not exceed the capabilities of pretrained models \cite{yue2025does} in domains with verifiable and structured answers. While NOVER focuses on free-form domains, we conduct a false premise task that similarly reveals the limits of RL. We adopt FANToM \cite{kim2023fantom}, a theory-of-mind dataset where models answer questions from a specific character’s perspective in multi-party dialogues. Some questions rest on false premises (e.g., asking for character A’s opinion on a movie when A was absent during the discussion on the movie in 
 the conversation), making them unanswerable. Experiments show that SFT can memorize such refusal patterns from training data, whereas NOVER depends on the pretrained model to generate candidate responses and selectively reinforce the better ones. Lacking exposure to refusal behavior during pretraining, the model struggles to reject false-premise questions, resulting in weaker performance than SFT (see \autoref{tab:fantom} and \autoref{appendix:false_premise}). These findings suggest that future work should consider integrating multiple post-training strategies.
 
\subsection{Curse of Proxy}
\label{sec:curse}
\begin{figure}[htbp]
    \centering
    \includegraphics[width=0.999\linewidth]{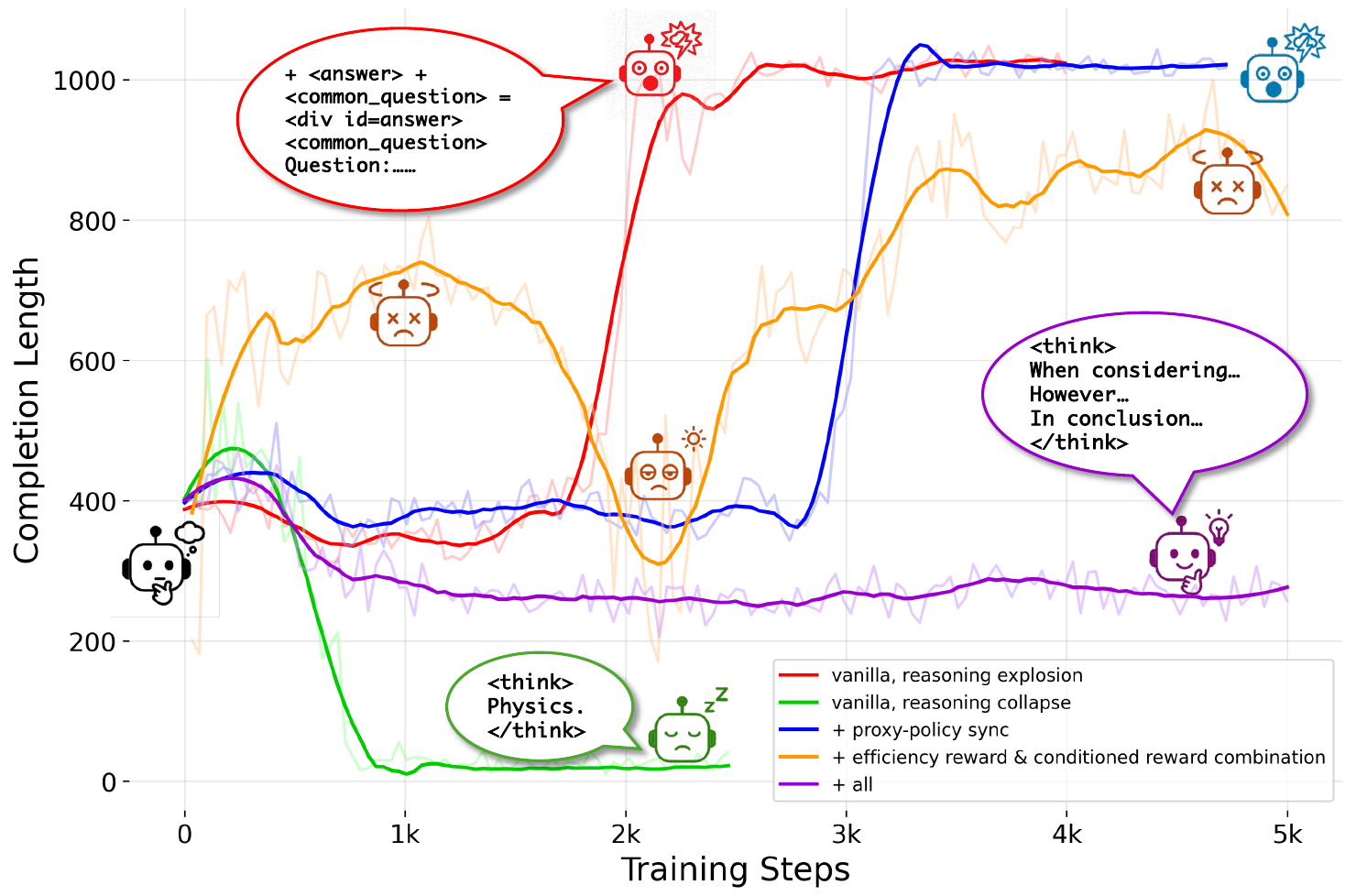}
    \caption{The curse of proxy: inaccurate proxy reward may lead to reward hacking.}
    \label{fig:curse}
\end{figure}

NOVER uses reasoning perplexity as a proxy for the ideal reward, a common practice in RL \cite{ng2000algorithms, christiano2017deep}. However, this introduces the \textbf{curse of proxy}: \emph{imprecise rewards can lead to reward hacking \cite{amodei2016concrete}, where models exploit flaws in the proxy rather than genuinely improving performance}. To analyze this, we perform ablations on NOVER and track completion lengths during training (\autoref{fig:curse}). \ul{(1) Training with only $R_{\mathrm{f}}$ and $R_{\mathrm{r}}$} reveals two failure modes: \textcolor{BrickRed}{\emph{reasoning explosion} (red)}, which is overlong, often garbled outputs, and \textcolor{ForestGreen}{\emph{reasoning collapse} (green)}, where models stop thinking and generate minimum reasoning tokens. Both stem from proxy misalignment with the evolving policy. \ul{(2) Adding proxy-policy synchronization} mitigates this (\textcolor{NavyBlue}{blue}), though eventual explosion still occurs. \ul{(3) Introducing the efficiency reward $R_{\mathrm{e}}$ and conditioned reward combination} enables partial self-recovery (\textcolor{Orange}{orange}), as invalid completions receive zero reward, encouraging resampling. Yet, recovery is inefficient. \ul{(4) Full NOVER integration} yields stable training (\textcolor{Violet}{purple}): the model stays “sober,” optimizing reasoning length only when it improves outcomes. Synchronization further reduces proxy bias, supporting robust optimization.

\subsection{Verifier, Free, or Not}

\begin{table}[htbp]
\centering
\small
\begin{tabular}{c|c|cc}
\toprule
Group                              & Method                 & 3B & 7B \\ \midrule
\multirow{3}{*}{Baselines}         & $Base$                   & 18.40\%    & 20.70\%    \\
                                   & $+\ CoT$                   & 24.20\%    & 29.20\%    \\
                                   & $+\ SFT$                   & 27.30\%    & 33.50\%    \\ \midrule
\multirow{3}{*}{\begin{tabular}[c]{@{}c@{}}Model \\ as Verifier\end{tabular}} & $+\ GV$          & 18.30\%    & 30.00\%    \\
                                   & $+\ LJ$          & 21.40\%    & 3.80\%     \\
                                   & $+\ LJ_S$ & --         & 21.60\%    \\ \midrule
Verifier-Free                      & $+\ NOVER$                 & 28.10\%    & 36.60\%    \\ \bottomrule
\end{tabular}
\caption{Experiments on WebInstruct. We compare LLM-as-a-Judge ($LJ$) and the officially released verifier model for WebInstruct, the general verifier $GV$.}
\label{tab:verifier_compare}
\end{table}

\label{sec:compare_llm_judge}
To evaluate NOVER on reasoning tasks with difficult verification, we compared it against alternative verifier designs: an LLM-as-a-judge ($LJ$) and a fine-tuned verifier model. Experiments were conducted on WebInstruct, which includes an official general verifier model ($GV$) \cite{generalreasoner}. For $LJ$, we used Gemini-2.0-flash with two prompt variants: a lenient “judge” prompt, and a stricter version ($LJ_s$) (aligned with our evaluation setup). As shown in \autoref{tab:verifier_compare}, model-based verifiers were highly unstable. With $LJ$, lenient prompts encouraged reward hacking, where $\pi_{\theta}$ generated vague but superficially valid outputs to elicit positive rewards (e.g., giving rough ideas instead of precise answers). In contrast, strict prompts yielded sparse rewards and unstable training, thus, the 3B model failed to train. The dedicated verifier also proved unreliable, often misled by the policy model. For example, the policy might only sketch initial steps and prompt the verifier to complete the calculation, causing the verifier to abandon judgment and instead solve the task, then assign an undeserved positive reward.

\subsection{Incentivized Reasoning Patterns}
\label{sec:pattern}

\begin{figure}[htbp]
    \centering
    \includegraphics[width=0.99\linewidth]{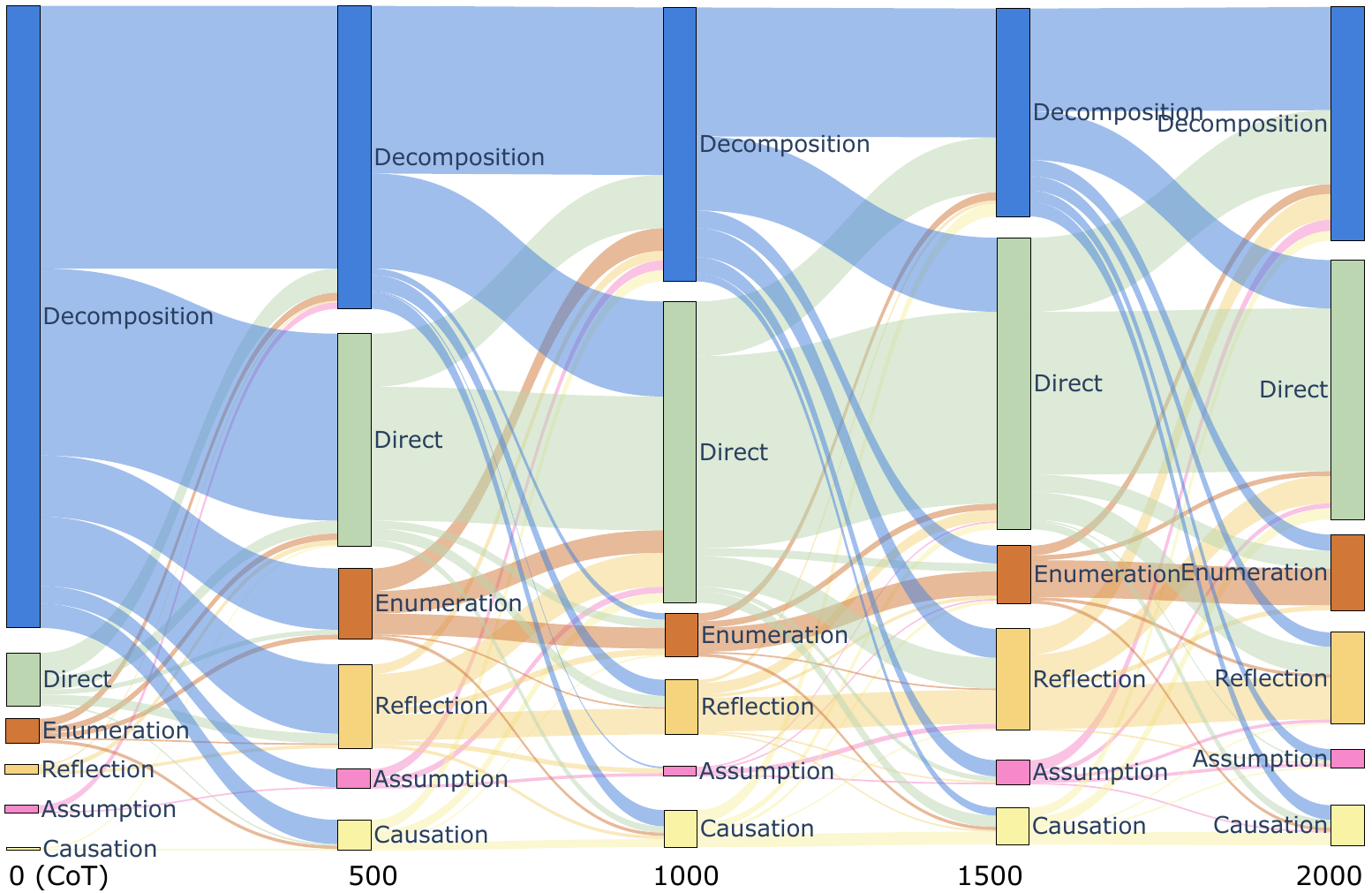}
    \caption{Change of reasoning patterns over steps.}
    \label{fig:pattern}
\end{figure}

We further investigated how reasoning patterns evolved during the training of NOVER. We extracted the reasoning tokens generated by the model at different training steps, including the outputs produced using CoT prompting before training (as 0 step), on a fixed test set. These reasoning traces were then classified using Gemini-2.0-flash into one of several predefined reasoning patterns. Following the design in~\citep{zeng2025simplerl}, we defined six types of reasoning patterns (see details in \autoref{appendix:pattern}). As shown in \autoref{fig:pattern}, before training, the model primarily used task decomposition typical of CoT prompting. As training progressed with the influence of the efficiency reward, the model learned to skip redundant reasoning steps when it could directly provide intermediate conclusions, resulting in a notable increase in the proportion of the direct reasoning pattern. At the same time, other reasoning types began to appear and gradually stabilized, reflecting the development of reasoning that is both effective and efficient.

\subsection{Inverse Incentive Training}\label{sec:rubric}
\begin{figure}[htbp]
    \centering
    \includegraphics[width=0.8\linewidth]{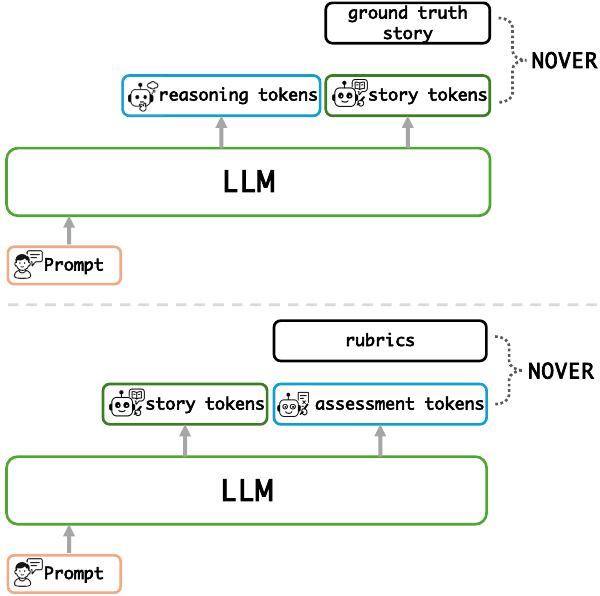}
    \caption{Comparison between standard NOVER training (up) and NOVER\_RUBRIC training (down).}
    \label{fig:rubric_paradigm}
\end{figure}
We further explore NOVER's flexibility in an interesting setting using the SS-GEN creative writing dataset \cite{feng2024ssgensocialstorygeneration}, where both ground truth stories and structural rubrics are provided.
\autoref{fig:rubric_paradigm} illustrates how we construct the inverse incentive training pipeline to enable rubric learning. 
Traditional incentive training requires the model to first generate thinking tokens, followed by answer tokens, with rewards computed to incentivize the intermediate thinking process. In the SS-GEN dataset’s story generation setup, by contrast, we require the model to generate story tokens first, followed by assessment tokens. There is no ground-truth story, but only ground-truth rubrics. In this setting, training with NOVER allows the intermediate tokens being incentivized, namely, the story tokens, to become the desired answer tokens.
We adapt NOVER by treating these rubrics as the guidelines for self-assessment, while the story becomes the intermediate process to incentivize, denoted NOVER\_RUBRIC.
This approach inverts the standard paradigm, implementing a “process as outcome” strategy. What we want is the intermediate tokens (story) instead of the final outputs (assessment). NOVER’s design thus effectively steers generation toward rubric satisfaction, without needing ideal story exemplars during training.
Empirical results show that Qwen2.5-7B’s accuracy improves from \textbf{50.79\%} (standard NOVER) to \textbf{64.37\%} after rubric-based training.

\begin{figure}[htbp]
    \centering
    \includegraphics[width=0.999\linewidth]{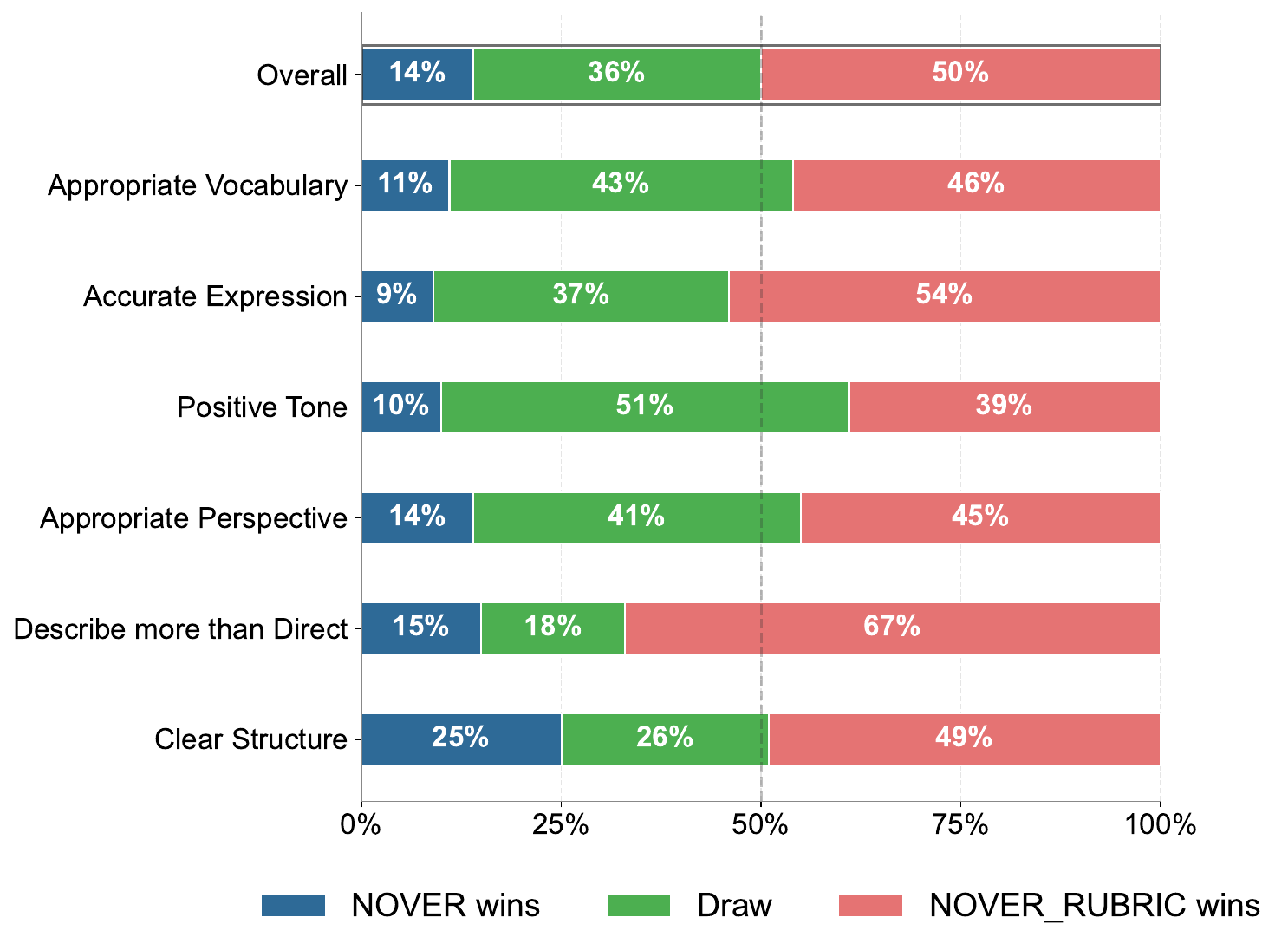}
    \caption{Human judgment on different rubrics for SS-GEN story generation.}
    \label{fig:rubric_result}
\end{figure}

To further validate the gains, we conducted human evaluations on each rubric dimension (\autoref{fig:rubric_result}), comparing stories from standard NOVER and the rubric-trained variant. The rubrics are detailed in \autoref{appendix:rubric}. Results consistently favor the latter across all criteria, highlighting NOVER’s flexibility for creative applications beyond classical reasoning tasks. It is worth noting that, due to the lack of explicit procedural rewards, relying solely on outcome-based rubric rewards can lead to occasional hallucinations in generated stories, e.g., repeating “once upon a time” twice. Future approaches will need to combine process and outcome rewards to advance the development of inverse incentive training.

\section{Conclusion}
In this paper, we present NOVER, a verifier-free framework for incentive training that pushes the boundaries of incentive learning to encompass any text-to-text task. NOVER demonstrates strong performance, and we discuss both the advantages and limitations of such a reinforcement learning paradigm, highlighting promising future directions for achieving robust and effective post-training. The flexibility of NOVER also enables novel learning paradigms such as inverse incentive training.

\section*{Limitations}

\citep{zeng2025simplerl, liu2025understanding} show that incentive training requires the base model to possess certain fine-tuned capabilities, such as partial CoT reasoning and instruction-following, to effectively incentivize reasoning. This observation is consistent with our findings on Qwen models. Moreover, general reasoning tasks in free-form format demand stronger base model capabilities than structured-answer tasks like math or coding. Currently, Qwen is a suitable choice in open-source models that meet the above requirements, so we conduct experiments on Qwen instead of other open-source models. Major open-source model teams have also recognized this and strengthened multi-stage pretraining by incorporating large amounts of CoT and instruction data, which gives us confidence that NOVER will be able to incentivize more models in the future.

\section*{Acknowledgments}
This work was supported in part by the UK Engineering and Physical Sciences Research Council through a Turing AI Fellowship (grant no. EP/V020579/1, EP/V020579/2) and the Prosperity Partnership scheme (grant no. UKRI566). Wei is supported by a PhD studentship provided by King's College London. The authors acknowledge the use of King’s Computational Research, Engineering and Technology Environment (CREATE) at King’s College London.

\bibliography{custom}

\clearpage
\appendix

\section{Dataset Construction Pipeline \& Statistics}
\label{appendix:dataset}
To minimize data contamination from pretraining, we prioritized the use of the most recent datasets available. We focused on general reasoning datasets that require multi-step inference and span multiple domains, not limited to STEM. In addition, we included text generation datasets that demand reasoning and planning abilities, such as
creative writing, social intelligence, and multilingual translation. All datasets were cleaned and filtered to ensure high-quality data. Specifically:

\begin{itemize}
  \item \textbf{Natural Reasoning} We excluded samples lacking a reference answer. We also filtered out samples where the reference answer was merely a number, a single word, or a single sentence, as such cases are often trivial for rule-based verifiers and do not reflect the open-ended reasoning tasks we aim to study.
  
  \item \textbf{General Thoughts and WebInstruct} These datasets underwent the same preprocessing as \emph{Natural Reasoning}. What's more, these two datasets contain multiple-choice questions, and we converted them into free-form QA format, ensuring that the reference answers included both the correct choice and its content. Due to inconsistent option formatting in the original data (such as 1., A), A, (a).), we designed seven regex-based patterns to clean and standardize the multiple-choice items.
  
  \item \textbf{EmoBench} We selected two subsets, \emph{emotional understanding} and \emph{emotional application}. For \emph{emotional understanding}, the prompt was adapted to specify a dual-option response format, reflecting the original structure in which each question is paired with two multiple-choice items (with four candidates each), targeting emotion recognition and causality, respectively.
  
  \item \textbf{FANToM} We chose two subsets aligned with free-form reasoning, which are \emph{full\_fact} (asking for an entity) and \emph{full\_belief\_gen} (asking for opinion in the perspective of somebody).
  
  \item \textbf{Other Datasets} No preprocessing was applied to the remaining datasets.
\end{itemize}

For datasets with existing train/test splits, we retained them. For others, we created splits by evenly sampling across subcategories (e.g., academic subjects) to ensure distributional consistency. For very large datasets, we sample 2{,}000 examples for training, 1{,}000 for validation, and 1{,}000 for testing. 

\begin{table}[htbp]
\centering
\tiny
\begin{tabular}{c|rrrrr}
\toprule
\textbf{Dataset} & \textbf{\#Original} & \textbf{\#Filtered} & \textbf{\#Train} & \textbf{\#Validation} & \textbf{\#Test} \\
\midrule
NR   & 1,150,000 & 192,178  & 2,000 & 1,000 & 1,000 \\
GT   & 431,000   & 78,483   & 2,000 & 1,000 & 1,000 \\
WI   & 232,000   & 231,833  & 2,000 & 1,000 & 1,000 \\
SGN  & 5,087     & 5,087    & 4,070 & 509   & 508   \\
EB   & 400       & 400      & 272   & --    & 128   \\
TB   & 2,860     & 2,860    & 1,432 & 571   & 857   \\
OPUS & 1,250,632 & 1,250,632 & 2,000 & 1,000 & 1,000 \\
FT   & 1,863     & 1,863    & 1,308 & --    & 555   \\
\bottomrule
\end{tabular}
\caption{Dataset statistics after filtering and splits into training, validation, and test sets.}
\label{tab:dataset_stats}
\end{table}
The detailed statistics are shown in \autoref{tab:dataset_stats}, where  \textbf{NR} is Natural Reasoning, \textbf{GT} is General Thoughts-430k, \textbf{WI} is WebInstruct, \textbf{SGN} is SS-GEN, \textbf{EB} is EmoBench, \textbf{TB} is TomBench, \textbf{OP} is OPUS-BOOK-TRANSLATION, \textbf{FT} is FANToM.

\section{Profiling}
\label{appendix:profiling}
\begin{figure}[htbp]
    \centering
    \includegraphics[width=0.999\linewidth]{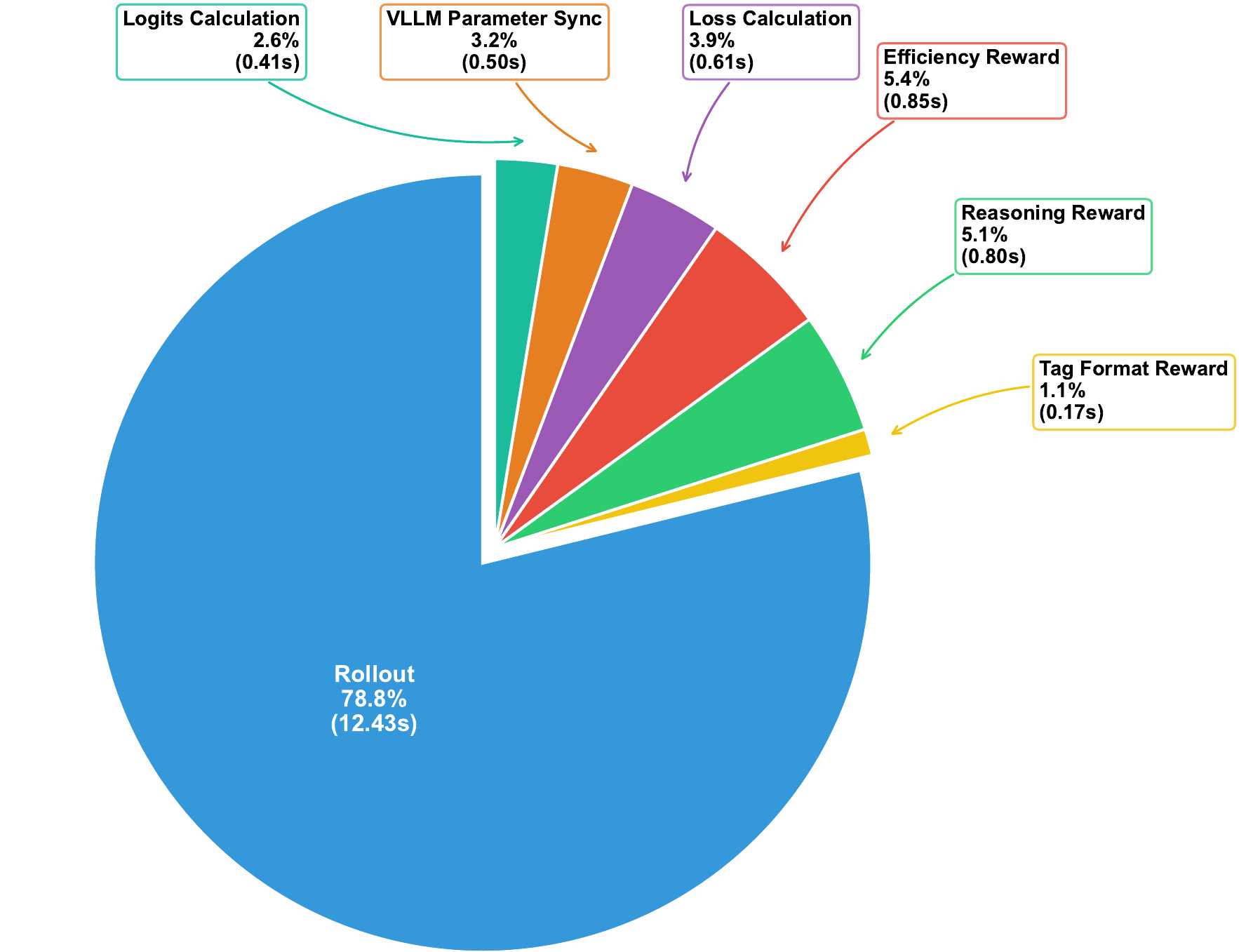}
    \caption{Profiling of the training time in NOVER.}
    \label{fig:profiling}
\end{figure}

\autoref{fig:profiling} shows the average training time per sample using two H100 GPUs with a completion length of 2,048 tokens. Most of the time is spent on the rollout stage, where vLLM \cite{kwon2023efficient} is used to efficiently sample multiple completions from the policy model. In contrast, computing the reasoning and efficiency rewards accounts for only about 5\% of the total training time.

\section{Hyperparameters}
\label{appendix:hyperparams}

\begin{table}[htbp]
\small
\centering
\begin{tabular}{ccc}
\toprule
Notation & Definition & Value \\
\midrule
$r$ & LoRA rank & 16 \\
$\alpha_{LoRA}$ & LoRA scaling factor & 32 \\
$p_{\text{dropout}}$ & LoRA dropout ratio & 0.1 \\
$dtype$ & Training precision & \texttt{bfloat16} \\
$G$ & Group Size & 8 \\
$B$ & Batch size per GPU & 8 \\
$\beta$ & KL coefficient & 0.1 \\
$w_f$ & Format reward weights & 1 \\
$w_r$ & Reasoning reward weights & 1 \\
$w_e$ & Efficiency reward weights & 1 \\
$\tau$ & Rollout temperature & 0.6 \\
$\eta_{\text{max}}$ & Maximum learning rate & $1\mathrm{e}{-5}$ \\
$\alpha$ & Synchronization coefficient & 0.9 \\
$T_{\text{sync}}$ & Synchronization interval & 100 \\
$L_{\text{gen}}$ & Max completion length & [512, 2048] \\
$\epsilon$ & clip range & 0.1 \\
$\epsilon_{\text{high}}$ & Upper clip range & 0.2 \\
$T_{max}$ & Max training steps & 5000 \\
\bottomrule
\end{tabular}
\caption{Hyperparameters used for NOVER training.}
\label{tab:hyperparams}
\end{table}

All hyperparameters for training NOVER are listed in \autoref{tab:hyperparams}. For general reasoning tasks and the SS-GEN creative writing task, we set the maximum completion length to 2048, while for other tasks, it is limited to 512. Training is run for up to 5,000 steps, although we typically apply early stopping when the validation reward, especially the tag format reward, no longer improves. For 3B models, training generally requires more steps than for 7B models.

\section{System Prompt}
\label{appendix:system_prompt}

\autoref{fig:system_prompt}, \autoref{fig:system_prompt_choice}, and \autoref{fig:system_prompt_translation} show the system prompt for three kinds of tasks in our experiments. We used minimal instructions in the prompt, relying solely on the \texttt{<think>}  and \texttt{<answer>} tags, without explicitly directing the model to reason. This approach minimizes external intervention and maximizes the model’s own incentivized capabilities. 

\begin{figure}[htbp]
\small
\begin{tcolorbox}[title={System Prompt for QA}]

Question: \{question\}

Answer the question and return in the following format:

\

<think>

...

</think>

<answer>

...

</answer>
\end{tcolorbox}
\caption{System prompt for most general reasoning tasks.}
\label{fig:system_prompt}
\end{figure}

\begin{figure}[htbp]
\small
\begin{tcolorbox}[title={System Prompt for Choice task}]

Question: \{question\}

Options: \{options\}

Answer the question and return in the following format:

\

<think>

...

</think>

<answer>

[option letter]: [option text]

</answer>
\end{tcolorbox}
\caption{System prompt for choice-format tasks.}
\label{fig:system_prompt_choice}
\end{figure}

\begin{figure}[htbp]
\small
\begin{tcolorbox}[title={System Prompt for Translation}]

Translate the following text 

from \{source\_lang\_name\} to \{target\_lang\_name\}:

\

\{source\_text\}

\

Answer the question and return in the following format:

\

<think>

...

</think>

<answer>

...

</answer>
\end{tcolorbox}
\caption{System prompt for translation tasks.}
\label{fig:system_prompt_translation}
\end{figure}

\section{LLM-as-Judge Prompt}
\label{appendix:judge_prompt}

\begin{figure}[htbp]
\small
\begin{tcolorbox}[title={Judge Prompt}]
Please determine if the following Model answer matches the Ground truth.

Only consider if the response matches the reference, regardless of how detailed or comprehensive it is.

Ground truth: \{reference\}

Model answer: \{response\}

Please provide your judgment in the following format:

\

<think>

[Your detailed reasoning about why the answers match or don't match]

</think>

<conclusion>

[yes/no]

</conclusion>

\

Return only the formatted response with reasoning and conclusion.
\end{tcolorbox}
\caption{Judge prompt (the strict version) used for evaluation and model-as-judge experiments (\Sref{sec:compare_llm_judge}).}
\label{fig:judge_prompt_strict}
\end{figure}

\begin{figure}[htbp]
\small
\begin{tcolorbox}[title={Judge Prompt, Lenient Version}]
Please determine if the following Model answer matches the Ground truth.

Does the model's answer broadly align with the reference answer, even if it’s not exactly the same?

Ground truth: \{reference\}

Model answer: \{response\}

Please provide your judgment in the following format:

\

<think>

[Your detailed reasoning about why the answers match or don't match]

</think>

<conclusion>

[yes/no]

</conclusion>

\

Return only the formatted response with reasoning and conclusion.
\end{tcolorbox}
\caption{Judge prompt (the lenient version) used in model-as-judge experiments (\Sref{sec:compare_llm_judge}).}
\label{fig:judge_prompt_lenient}
\end{figure}

\autoref{fig:judge_prompt_strict}, \autoref{fig:judge_prompt_lenient}, and \autoref{fig:judge_prompt_gv} show the judgment prompt used in evaluation and model-as-judge experiments in \Sref{sec:compare_llm_judge}. The judge prompt for the general verifier is taken from the official Huggingface repository~\footnote{\url{https://huggingface.co/TIGER-Lab/general-verifier}}. We also follow the generation configuration on the official repository, where the temperature is set to 0.0 and the maximum tokens to 1,024.

\begin{figure}[htbp]
\small
\begin{tcolorbox}[title={Judge Prompt, General Verifier}]
User: \#\#\# Question: \{question\}

\#\#\# Ground Truth Answer: \{reference\_answer\}

\#\#\# Student Answer: \{answer\_content\}

For the above question, please verify if the student's answer is equivalent to the ground truth answer.

Do not solve the question by yourself; just check if the student's answer is equivalent to the ground truth answer.

If the student's answer is correct, output "Final Decision: Yes". If the student's answer is incorrect, output "Final Decision: No". Assistant:"

\end{tcolorbox}
\caption{Judge prompt from the official code of general verifier \cite{generalreasoner}.}
\label{fig:judge_prompt_gv}
\end{figure}

\section{Reasoning Hallucination Case}
\label{appendix:case}

\begin{figure*}[htbp]
\small
\begin{tcolorbox}[title={Case Study in OPUS Translation Task}]

\textbf{Task:}

Translate the following text from English to Spanish:

With her perfect figure and elegant dress she was, indeed, a strange apparition upon a lonely moorland path.

\rule[3pt]{\textwidth}{0.20mm}

\textbf{Reference:}

Dada la perfección y elegancia de su vestido, resultaba, desde luego, una extraña aparición en la solitaria senda del páramo.

\rule[3pt]{\textwidth}{0.20mm}

\textbf{Base Model CoT:}

To translate the given text from English to Spanish, I will follow these steps:

1. Identify the main components of the sentence: subject, verb, and object.

2. Translate each component individually.

3. Rearrange the components to form a grammatically correct sentence in Spanish.

4. Ensure that the translated sentence maintains the original meaning and tone.

Step 1: Identify the main components of the sentence.

Subject: "She"

......

Step 2: Translate each component individually.

Subject: "Ella" (She)

......

Step 3: Rearrange the components to form a grammatically correct sentence in Spanish.

"Ella era, en efecto, una extraña aparición en un sendero solitario de tierra mojada."

Step 4: Ensure that the translated sentence maintains the original meaning and tone.

The translated sentence maintains the original meaning and tone, \colorbox{pink}{\textit{as it describes the woman's appearance and the setting}}

\rule[3pt]{\textwidth}{0.20mm}

\textbf{Base Model Translation:}

Ella era, en efecto, una extraña aparición en un sendero solitario de tierra mojada.

\rule[3pt]{\textwidth}{0.20mm}

\textbf{NOVER thinking process:}

<think>

To translate the given text from English to Spanish, I will follow these steps:

1. Identify the main components of the sentence: subject, verb, object, \colorbox{yellow}{\textit{and any additional descriptive elements.}}

2. Translate each component into Spanish while maintaining the original meaning and context.

3. Rearrange the translated components to fit the Spanish sentence structure, \colorbox{yellow}{\textit{which may differ from English.}}

4. Review the final translation for accuracy and fluency.

Here's the step-by-step translation process:

1. Main components:

   - Subject: "she"

......

   - \colorbox{yellow}{\textit{Additional descriptive elements: "perfect figure", "elegant dress", "indeed", "upon a lonely moorland path"}}

2. Translations:

   - "she" -> "ella"

......

   \colorbox{yellow}{\textit{- "perfect figure" -> "figura perfecta"}}

   \colorbox{yellow}{\textit{- "elegant dress" -> "traje elegante"}}

   - "indeed" -> "realmente"

   - "upon a lonely moorland path" -> "en una solitaria senda de moorland"

4. Reviewing the final translation:

   - The translation maintains the original meaning and context, and the sentence structure is appropriate for Spanish.

I have translated the given text from English to Spanish by identifying the main components, translating each component, rearranging them to fit the Spanish sentence structure, and reviewing the final translation for accuracy and fluency.

</think>

\rule[3pt]{\textwidth}{0.20mm}

\textbf{NOVER translation:}

<answer>

Ella era, realmente, una extraña aparición en una solitaria senda de moorland con su figura perfecta y traje elegante.

</answer>

\end{tcolorbox}
\caption{A case in the OPUS translation dataset, where the base model generates hallucinated CoT that leads to wrong translation, while NOVER follows similar steps but corrects the wrong parts in the thinking process.}
\label{fig:opus_case}
\end{figure*}

\autoref{fig:opus_case} presents an example from the OPUS translation dataset where both the base model and NOVER perform reasoning to generate a translation. As shown, the base model is able to produce seemingly correct reasoning steps, sometimes nearly identical to those of NOVER. However, despite generating the right steps, the base model fails to reach the correct conclusion. For instance, its reasoning chain ends with a self-check that asserts the translation is perfect and faithfully conveys the meaning of the original sentence, including the description of the “woman’s appearance.” Yet, in the earlier steps, the model translates only the subject, verb, and object, omitting any reference to the woman’s appearance. This kind of confident but incorrect self-verification, known as one kind of reasoning hallucination \cite{huang2025survey, qi2024can}, can cause the model to overlook essential content, resulting in worse performance than directly producing the translation. In contrast, NOVER follows a similar reasoning path but incorporates critical refinements. For example, it explicitly notes the need to translate not only the core grammatical elements but also any additional descriptive components. As a result, in the initial step of identifying main elements, NOVER successfully detects phrases related to the woman’s appearance, such as “perfect figure” and “elegant dress,” and accurately translates and integrates them in later steps. Moreover, NOVER provides more rigorous rubrics, such as emphasizing in the rearrangement step that Spanish word order “may differ from English.”

\section{Reasoning Perplexity Example}
\label{appendix:RP_case}

\begin{figure*}[htbp]
\small
\begin{tcolorbox}[title={Two Completions of NOVER on Natural Reasoning Dataset}]

\textbf{Question:}

What is the primary mechanism by which the body eliminates excess cholesterol, and how do dietary factors such as fiber intake and fatty food consumption affect this process? Provide a detailed explanation of the gastrohepatic cycle and the role of bile secretion in cholesterol elimination.

\rule[3pt]{\textwidth}{0.20mm}

\textbf{Reference:}

The primary mechanism by which the body eliminates excess cholesterol is through the gastrohepatic cycle, where excess cholesterol is secreted by the bile ducts back into the gastrointestinal tract. Dietary factors such as fiber intake can help prevent the digestion of cholesterol, while fatty food consumption can increase cholesterol production.

\rule[3pt]{\textwidth}{0.20mm}

\textbf{NOVER completion 1, with reasoning perplexity = 1.406 (the good one)}

<think> 

The primary mechanism by which the body eliminates excess cholesterol involves a process called the gastrohepatic cycle. This cycle consists of several steps:

1. Cholesterol in the blood is taken up by the small intestine and converted into bile acids, which are then secreted into the intestine.

2. Bile acids are reabsorbed by the small intestine and returned to the liver via the portal vein.

3. In the liver, the bile acids are converted back into cholesterol and stored or used for other purposes.

4. When the body has an excess of cholesterol, the liver produces more bile acids, which are excreted in the feces.

......

In summary, the gastrohepatic cycle is the primary mechanism by which the body eliminates excess cholesterol, and dietary factors such as fiber intake and fatty food consumption can affect this process in various ways.

</think>

<answer>

The primary mechanism by which the body eliminates excess cholesterol involves a process called the gastrohepatic cycle. This cycle consists of several steps: Cholesterol in the blood is taken up by the small intestine and converted into bile acids, which are then secreted into the intestine. Bile acids are reabsorbed by the small intestine and returned to the liver via the portal vein. In the liver, the bile acids are converted back into cholesterol and stored or used for other purposes. When the body has an excess of cholesterol, the liver produces more bile acids, which are excreted in the feces. Dietary factors such as fiber intake and fatty food consumption can affect this process in various ways. Dietary fiber, particularly soluble fiber, can bind to bile acids in the intestine and prevent their reabsorption. This leads to an increased excretion of bile acids in the feces, which in turn increases the amount of cholesterol that is eliminated from the body. Fatty foods, particularly those high in saturated and trans fats, can increase the amount of cholesterol produced by the liver. This can lead to an increase in the amount of cholesterol that needs to be eliminated from the body. Certain fatty acids, such as omega-3 fatty acids, can also have a beneficial effect on cholesterol metabolism. Omega-3 fatty acids can reduce the amount of cholesterol produced by the liver and increase the amount of cholesterol that is eliminated from the body.

</answer>

\rule[3pt]{\textwidth}{0.20mm}

\textbf{NOVER completion 2, with reasoning perplexity = 1.516 (the bad one)}

<think>

The primary mechanism by which the body eliminates excess cholesterol is through the excretion of cholesterol-rich bile acids in the feces. This process involves the breakdown and conversion of cholesterol into bile acids in the liver, followed by the release of these bile acids into the small intestine during digestion. Bile acids are then excreted from the body in the feces.

Dietary factors such as fiber intake and fatty food consumption can affect this process in several ways. Fiber-rich foods can increase the viscosity of the stool, which slows down the passage of bile acids through the digestive tract. This allows more time for the bile acids to be absorbed back into the bloodstream and reused, reducing the amount of cholesterol that is excreted in the feces.

......

Bile secretion plays a critical role in cholesterol elimination by providing the necessary environment for the breakdown and conversion of cholesterol into bile acids in the liver. Bile acids are then released into the small intestine during digestion, where they interact with fatty acids and triglycerides to form micelles. These micelles help to emulsify the fatty acids and make them more available for absorption by the small intestine.

</think>

<answer>

The primary mechanism by which the body eliminates excess cholesterol is through the excretion of cholesterol-rich bile acids in the feces. Dietary factors such as fiber intake and fatty food consumption can affect this process by increasing or decreasing the amount of cholesterol that is excreted in the feces. The gastrohepatic cycle is the process by which bile is produced in the liver, stored in the gallbladder, and released into the small intestine during digestion. Bile secretion plays a critical role in cholesterol elimination by providing the necessary environment for the breakdown and conversion of cholesterol into bile acids in the liver.

</answer>

\end{tcolorbox}
\caption{We show two completions from NOVER on the Natural Reasoning dataset, showing how reasoning perplexity discriminates good responses from bad ones.}
\label{fig:nr_case}
\end{figure*}

\autoref{fig:nr_case} shows two completions generated by NOVER in response to a medical question from the natural reasoning dataset, each associated with a different level of reasoning perplexity. As illustrated, Completion 1, which has lower reasoning perplexity, is clearly superior to Completion 2 in terms of physiological mechanisms, terminological accuracy, and the discussion on “the role of dietary factors”, and overall scientific reasoning. In contrast, Completion 2 is relatively coarse and contains misleading or missing descriptions of relevant terms and mechanisms. Although reasoning perplexity does not directly compare the answer with a reference, it estimates the likelihood of the reference based on the inferred reasoning path, thereby providing a concise and effective measure of answer quality, as evidenced by this example.

\section{False Premise Example}
\label{appendix:false_premise}
\autoref{fig:fantom_case} illustrates an example from the FANToM dataset, shedding light on why NOVER may underperform compared to SFT in the false premise task setting. As shown, the example involves a multi-party dialogue and poses a question requiring higher-order theory of mind reasoning—for instance, inferring how person B would think about a certain topic from person A’s perspective. The trap, however, is that B may not have participated in the conversation and thus is unaware of the topic, making the question unanswerable. The corresponding reference typically reflects this, such as “B is unaware of…”. SFT can quickly memorize and reproduce such response patterns (e.g., “somebody is unaware of…”), whereas NOVER relies on the base model to first generate a correct answer, which is then reinforced. However, the base model may struggle to recognize the false premise and might not even consider the question potentially unanswerable. Consequently, as shown in the figure, NOVER attempts to answer the question but remains constrained by the flawed assumption that B is aware of the topic, and reasons accordingly. This highlights why NOVER may perform worse than SFT on counterintuitive tasks. Addressing such cases may require a hybrid approach: first using SFT to expose the model to appropriate response patterns and challenge its assumptions, followed by reinforcement learning to strengthen accurate reasoning.

\begin{figure}[htbp]
\small
\begin{tcolorbox}[title={A False Premise Example in FANToM Dataset}]

\textbf{Here is a dialogue:}

\begin{itemize}
    \item Eliana: Hello, guys. I find it fascinating how various cultures influence food and cuisine. For instance, in my culture, Brazilian cuisine is a culmination of Europe, Africa, and Native American influences.
    \item Kade: How so, Eliana? Could you give an example?
    \item Eliana: Absolutely, Kade! Take feijoada for example, it's a Brazilian dish that's similar to a bean stew, filled with beef and pork. It's traditionally served with rice, collard greens, and farofa, a cassava flour mixture. Feijoada was often made by slaves, who would use whatever meats were available, hence the mixture.
    \item ......
    \item Kade: Definitely, Alberto. So, I guess we meet at the next food tasting session, where we have a taste of Brazil, Scotland, Canada, and Mexico!
\end{itemize}

\

\textbf{Question:}

\

What dishes from Scotland and Canada does Eliana think that Alberto believes were discussed by Addison and Kade respectively?

\rule[3pt]{\textwidth}{0.20mm}

\textbf{Reference:} 

\

\colorbox{yellow}{\textit{Alberto is unaware about}} which dishes from Scotland and Canada were discussed by Addison and Kade respectively as he was not involved in the conversation when this was discussed. Thus, Eliana does not have any basis to form an opinion on what Alberto believes about the mentioned dishes.

\rule[3pt]{\textwidth}{0.20mm}

\textbf{NOVER Response:}

\

<think>

Addison and Kade were the ones who discussed the dishes from Scotland and Canada. From the dialogue, it is evident that Addison mentioned the dish "haggis", which is a combination of sheep's heart, liver, and lungs minced with onion, oatmeal, suet, and spices......
So, Alberto believed that Addison and Kade discussed the dishes from Scotland and Canada, which are haggis and poutine respectively.

</think>

<answer>

Haggis is a dish from Scotland. Kade mentioned the dish "powwow" during the conversation, which is a traditional Mexican festival. So, \colorbox{pink}{\textit{Alberto believed that}} Addison and Kade discussed the dishes from Scotland and Canada, which are haggis and poutine respectively.

</answer>

\end{tcolorbox}
\caption{A false premise example in FANToM \cite{kim2023fantom} dataset, where the question cannot be answered.}
\label{fig:fantom_case}
\end{figure}

\section{Rubrics in Inverse Incentive Training}
\label{appendix:rubric}
To evaluate whether this inverse incentive training approach (treating the process as the desired outcome) successfully aligns the generated story tokens with the rubrics targeted by the final assessment tokens, we developed a human evaluation platform. We sampled 100 examples and asked human annotators to express preferences for each story concerning each rubric. The details of each rubric are provided to human annotators, including:
\begin{enumerate}
    \item \textbf{Clear Structure} A clear structure refers to that there is a straightforward title establishing the intervention goal, an introduction explaining the story theme and setting, a main body depicting the specific situation and appropriate responses, and a conclusion summarizing the core information presented.
    \item \textbf{Describe more than Direct} The number of descriptive sentences should exceed twice the number of coaching sentences in a Social Story. A descriptive sentence accurately describes the interaction, an event, or an explanation of the rationale that underlies what and why people think, say, or do, including perspective and affirmative sentences. Coaching sentences gently and safely guide the behavior.
    \item \textbf{Appropriate Perspective }A): The Social Story should never use the second-person perspective to prevent being overly direct. B): When describing negative behaviors, the Social Story should never employ the first-person perspective to safeguard the dignity and esteem of the audience.                                    
    \item \textbf{Positive Tone} The Social Story should always describe situations or guide behaviors in a positive manner, especially when depicting typical or expected behaviors in specific situations.
    \item \textbf{Accurate Expression} The story should use language that is as clear and unambiguous as possible because ASD children typically interpret things literally rather than inferring implicit meanings that require social insight.
    \item \textbf{Appropriate Vocabulary} The Social Story should choose the most comfortable and accurate vocabulary for the audience. Firstly, use positive verbs while also being mindful of the varying implications of verbs. Avoid using terms that are likely to evoke strong emotional feelings such as "shouldn't", "must", "supposed to" and so on.          
\end{enumerate}

\section{Reasoning Patterns}
\label{appendix:pattern}
We categorize reasoning patterns in the NOVER incentivized reasoning process into six main kinds. Here is the definition of each kind:
\begin{enumerate}
    \item Direct: The direct recall of factual information, definitions, or established concepts without further analysis or transformation. This pattern involves stating information from memory as-is.
    \item Decomposition: The systematic organization of a problem into manageable components, establishing clear steps, intermediate goals, or methodical frameworks. This pattern involves creating a structured approach to solving complex problems.
    \item Enumeration: The listing of multiple possibilities, options, alternatives, or cases without immediately selecting or committing to any specific one. This pattern involves comprehensively covering various aspects or potential scenarios.
    \item Reflection: The process of revisiting, questioning, or reassessing previously stated ideas, assumptions, or conclusions. This pattern involves reflecting on one's own reasoning and making adjustments based on further consideration.
    \item Assumption: The introduction of hypothetical conditions or premises that serve as a foundation for further reasoning. This pattern involves making conditional statements to explore potential scenarios or outcomes.
    \item Causation: The establishment of cause-effect relationships between events, actions, or conditions. This pattern involves explaining how one factor leads to or influences another.
\end{enumerate}

\end{document}